%% file: arxiv.tex
\documentclass[letterpaper, 10 pt, conference]{ieeeconf}  % Comment this line out if you need a4paper

\usepackage{color-edits}
\usepackage[dvipsnames]{xcolor}
\definecolor{cmured1}{HTML}{C41230}
\definecolor{cmured2}{HTML}{941120}
\title{\LARGE \bf
ActiveScale: Scaling Active Perception for Robots across 
\\
Model, Data, and Hardware
}

\author{Shuai Zhou$^{1,*,\dagger}$, Kaisheng Pang$^{2,*}$, Wenxuan Song$^{2,\ddagger}$, Wenjie Zhang$^{2}$, Xinhu Zheng$^{2}$, Haoang Li$^{2, \ddagger}$% <-this % stops a space
\vspace{1mm}\\
$^{1}$Robotics Institute, Carnegie Mellon University \\ $^{2}$The Hong Kong University of Science and Technology (Guangzhou) \vspace{1mm}\\
*Equal contribution, $\dagger$Project lead, $\ddagger$Equal advising
}

\usepackage{multirow}
\usepackage{booktabs}
\usepackage{subcaption}
\usepackage{algorithm}
\usepackage{algorithmic}

\usepackage{pifont}
\usepackage{hyperref}
\usepackage[dvipsnames]{xcolor}
\usepackage{graphicx}
\usepackage{float}
\usepackage{amsmath}
\usepackage{adjustbox}
\usepackage{array}
\usepackage{amsfonts}
\usepackage{amssymb}
\usepackage{bm}
\usepackage{cite}
\usepackage[font=small]{caption}

\let\labelindent\relax
\usepackage[inline]{enumitem}
\newcommand{\method}{ActiveScale}
\hypersetup{
    colorlinks=true,
    pdfborder={0 0 0},
    urlcolor=cmured1
}

\begin{document}

\thispagestyle{empty}
\pagestyle{empty}

% \maketitle
\twocolumn[{
\renewcommand\twocolumn[1][]{#1}
\maketitle
\vspace{-0.23in}
\begin{center}
\noindent\begin{minipage}{1.0\textwidth}
    \includegraphics[width=1.0\linewidth]{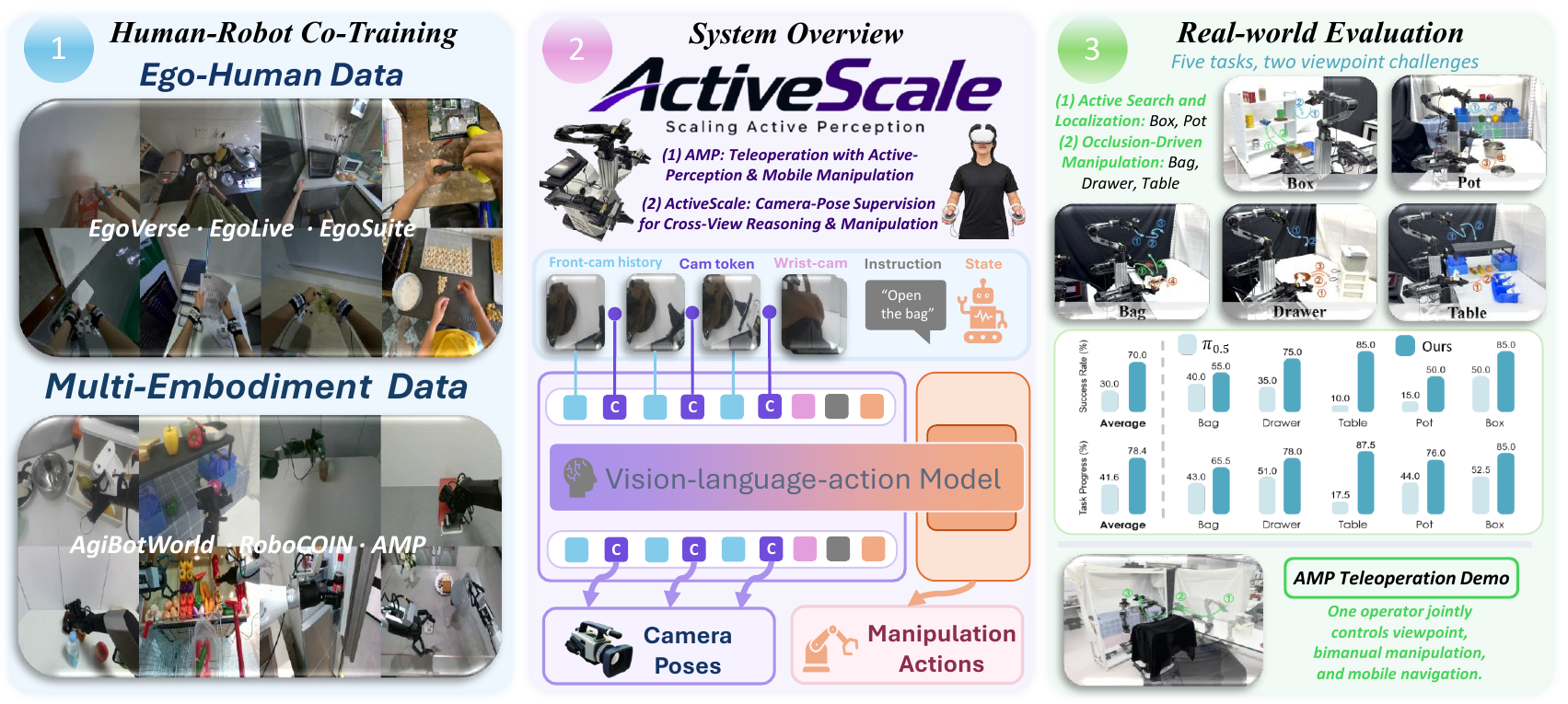}
    \end{minipage}\hspace{0.05in}
    \vspace{-0.1cm}
    \captionof{figure}{
    Overview of \method: \textbf{(1)} egocentric human data provide a scalable path for transferring active-perception priors to robots through human--robot co-training, \textbf{(2)} Active-perception Mobile-manipulation Platform (AMP) enables coordinated active-perception and mobile-manipulation teleoperation, while pose-grounded temporal modeling supports cross-view reasoning and action generation, and \textbf{(3)} real-world evaluation assesses these capabilities across five manipulation tasks and a mobile-manipulation demonstration.
    }
    \label{fig:head}
    \vspace{-0.23cm}
\end{center}
}]

%%%%%%%%%%%%%%%%%%%%%%%%%%%%%%%%%%%%%%%%%%%%%%%%%%%%%%%%%%%%%%%%%%%%%%%%%%%%%%%%
\begin{abstract}
% 在固定视角无法充分揭示任务相关信息、且存在遮挡或未观测区域时，主动感知对于机器人操作至关重要。然而，使视觉-语言-动作（VLA）模型能够理解不断变化的视角，并主动获取有价值的观测信息，仍然面临挑战。我们提出了 \method，一个从模型、数据和硬件协同推进主动感知的框架。

% 在模型方面，我们为 VLA 引入历史视频观测和显式相机位姿监督：通过为每一帧加入位姿 token，并使用轻量级预测头，模型能够关联不同视角下的观测，从而形成对场景的一致理解。在数据方面，为了利用人类第一视角数据中天然包含的相机运动信号，我们在超过 3,000 小时的第一视角视频上对模型进行中期训练，使其适应时序输入和位姿监督，并学习相机运动。在硬件方面，我们进一步提出 AMP：该平台支持主动感知与移动操作，并允许单个操作者同步完成二者的遥操作，从而为需要协调视角变化与操作的任务提供高效的数据采集能力。

% 实验表明，所提出的方法能够提高主动感知任务的成功率；消融实验进一步验证了相机位姿感知建模与第一视角中期训练的有效性。整体而言，这些设计为研究和发展机器人操作中的主动感知提供了一个一体化基础。
Active perception is essential for robotic manipulation when fixed viewpoints leave task-relevant information occluded or unobserved. However, enabling vision-language-action (VLA) models to reason across changing viewpoints and actively acquire informative observations remains challenging. We present \method, a framework that advances active perception through coordinated model, data, and hardware designs. Our model augments a VLA with historical video observations and explicit camera-pose supervision, using per-frame pose tokens and a lightweight prediction head to associate observations across viewpoints and support a coherent understanding of the scene. To learn from the camera motion naturally present in human activity, we introduce a scalable human--robot mid-training recipe using 1000 hours of egocentric and robotic data, adapting the model to temporal inputs and pose supervision. We further introduce Active-perception Mobile-manipulation Platform (AMP), a robotic platform that supports active perception and mobile manipulation through single-operator teleoperation, enabling scalable collection of demonstrations that coordinate viewpoint changes and manipulation. Experiments demonstrate improved success rates on active-perception tasks, while ablation studies validate the contributions of camera-pose-aware modeling and egocentric mid-training. Together, these components provide an integrated foundation for studying and developing active perception in robotic manipulation. More information and
materials are available at \href{https://active-scale.github.io/}
{\textcolor{cmured1}{\texttt{active-scale.github.io}}}.
\end{abstract}

% \begin{figure*}[t]
%     \centering
%     \includegraphics[width=\textwidth]{figures/teaser.pdf}
%     \caption{
%     Overview of our proposed work.
%     }
%     \label{fig:tasks}
% \end{figure*}
%===============================================================================

\section{Introduction}
\label{sec:introduction}
	\input{contents/introduction}

\section{Related Work}
\label{sec:related_work}
	\input{contents/related_work}

\section{Methodology}
\label{sec:methodology}
	\input{contents/methodology}

\section{Experiments}
\label{sec:experiments}
	\input{contents/experiments}

\section{Conclusions}
\label{sec:conclusions}
	\input{contents/conclusions}

%                                   % the textheight of the last page by a suitable amount.
%                                   % This command does not take effect until the next page
%                                   % so it should come on the page before the last. Make
%                                   % sure that you do not shorten the textheight too much.

%%%%%%%%%%%%%%%%%%%%%%%%%%%%%%%%%%%%%%%%%%%%%%%%%%%%%%%%%%%%%%%%%%%%%%%%%%%%%%%%

% \section*{ACKNOWLEDGMENT}

%%%%%%%%%%%%%%%%%%%%%%%%%%%%%%%%%%%%%%%%%%%%%%%%%%%%%%%%%%%%%%%%%%%%%%%%%%%%%%%%

\bibliographystyle{IEEEtran}
\bibliography{references}

% \clearpage

% \label{sec:appendix}
% 	\input{contents/appendix}

\end{document}

%% file: contents/introduction.tex
% TBD

% what's the contribution? 
% An Active Perception System?
% A model? A scaling pattern with egocentic human videos?

% wenxuan:
% todo:
% contribution:
% 1.model
% 2.midtraining ready (model scaling)
% 3.hardware system(optional)
% 4.experiment()

%机器人基础模型，如VLA，能力的增强，使其能够在不同场景下完成不同任务。随着机器人场景更丰富，任务更复杂，固定的视角往往导致完成任务所需的观察是patially observed，这要求机器人在杂乱场景中有策略地调整观察视角，以获取与任务相关的信息，例如转动自我中心视角，从而看清隐藏在柜子中的碗、书包里的东西、减少遮挡。因此静态感知逐渐无法满足要求，主动感知越来越重要。
Recent advances in robotic foundation models, such as vision-language-action models (VLAs)~\cite{intelligence2025pi_, generalist2026gen1, team2026xiaomi, song2025pd}, have substantially broadened robots’ ability to perform diverse tasks across environments. However, as robotic settings become richer and tasks more complex, observations from a fixed viewpoint are often only partially informative. Successfully completing a task may require the robot to actively acquire task-relevant information in cluttered environments—for example, by reorienting its egocentric view to inspect a bowl hidden inside a cabinet or the contents of a backpack, thereby reducing occlusion. Consequently, static perception is increasingly insufficient, making \textbf{active perception} a critical capability for general-purpose robotic agents.

%先前的工作尝试解决主动感知的问题，sapave通过将相机运动和机械臂运动的训练过程解耦来实现对二者能力的良好学习；activemimic利用egocentric data进行active perception预训练；activeumi提出active perception的硬件解决方案。
%然而仍面临三个关键问题：1. 大规模static camera的robot data上预训练的vla直接在动态相机图像上训练时，难以理解视角的变换。2. egocentric data自然的带有camera motion，而现有方法无法有效利用ego data进行训练。3. 现有主动感知仍然局限于静态场景下的感知，因此和真实世界中的移动操纵中的主动感知存在gap。
% 9.11:这一段得润色一下或者直接删掉
% 9.13:润色好了
Prior work has explored several directions toward active perception. SAPAVE~\cite{liu2026sapave} decouples the training of camera motion and arm motion, enabling effective learning of both capabilities. ActiveMimic~\cite{lin2026activemimic} leverages egocentric data for active-perception pretraining, while EgoMI~\cite{yu2025egomi} introduces a hardware platform for scaling egocentric demonstration collection with approximately 200 hours of data for policy training.
Despite this progress, three key challenges remain. First, accomplishing key active-perception tasks requires a model to infer its camera pose and predict subsequent camera motion, with the latter contingent on the former. However, existing VLAs lack these capabilities. 
Second, although egocentric data naturally contains camera motion, existing methods do not effectively exploit this signal for learning active perception. 
Third, current approaches predominantly study active perception in static environments~\cite{liu2026sapave, lin2026activemimic}, leaving a substantial gap to real-world mobile manipulation, where both the robot and its observation viewpoint must adapt to dynamic, cluttered settings.
\textit{More importantly, a unified infrastructure for systematically studying active perception across model, data, and hardware remains lacking.}

%为了解决上述问题，我们提出了\method. 具体来说: 
%为了实现对一个coherent 3D world的感知，我们首先将输入的单帧图片改为带有历史帧的一段视频。接着，我们加入一个简单的任务：one camera pose token, instantiated from one of two learnable queries, is appended to each frame’s visual features, and a lightweight projector and head regress pose parameters from the VLM’s hidden states.这一设计不仅让VLA显式的理解了当前帧camera pose，并进一步将不同帧之间的视频观测关联起来，使得只需一组连续视频帧的输入即可perceive the persistent scene across viewpoints。、
To address the aforementioned challenges, we propose \textbf{\method}. It consists of 3 important designs:
(1) \textbf{Model.} To enable perception of a coherent 3D world, we first extend the input from a single image to a video sequence containing historical frames. 
We then introduce a simple task: 
%这一句从cambrain-p里抄过来没修改
% one camera pose token, instantiated from one of two learnable queries, is appended to each frame’s visual features, and a lightweight projector and head regress pose parameters from the VLM’s hidden states.
a learnable camera token follows each front-camera frame’s visual features, and a linear projector and auxiliary camera head predict translation, rotation, and field of view from the corresponding VLM hidden states.
This design enables the VLA not only to explicitly understand the camera pose of the current frame, but also to associate video observations across different frames, thereby allowing it to perceive the persistent scene across viewpoints using only a sequence of consecutive video frames as input.
%(2)数据上，为了充分训练模型学习camera motion的能力，我们在大规模egocentric data对(1)中所述的模型进行mid-train，数据总量超过3k小时。这个过程使得预训练模型充分适应增加的输入和监督，并更好的理解camera motion。
%(3)硬件上，我们提出AMP,他将双臂的设备扩展至支持主动感知和移动操作，并且可由一个操作员同时完成上述任务，大大提升了数据采集效率。
(2) \textbf{Data and mid-training.} To adequately train the model to learn camera motion, we mid-train the model described in (1) on large-scale egocentric data and robot data, totaling over 1000 hours. This provides a scalable training recipe that adapts the pretrained model to the augmented inputs and supervision while improving its understanding of camera motion.
(3) \textbf{Hardware.} We develop an Active-perception Mobile-manipulation Platform (\textbf{AMP})
that integrates bimanual manipulation, an independently actuated camera, and
mobile-base control through a unified Meta Quest 2 teleoperation interface.
A single operator can jointly control viewpoint, manipulation, and base motion,
providing a scalable interface for collecting coordinated mobile-manipulation demonstrations.

%大量实验证明我们提出的模型在主动感知任务上能够达到更高的成功率，。同时，一系列实验证明了我们的模型设计和mid-training的有效性。
Diverse and sufficient experiments demonstrate that our proposed model achieves higher success rates on active perception tasks. Moreover, ablation studies validate the effectiveness of our model design and mid-training procedure.
In summary, our contributions are threefold:
\begin{itemize}
    \item We propose a pose-grounded temporal VLA for active perception and
cross-view reasoning, achieving state-of-the-art performance across
five real-world active-perception tasks.
    \item We propose a human--robot mid-training recipe using large-scale egocentric human and robot data. To the best of our knowledge, it is the first approach to scale active-perception training to 1000 hours of data--more than 5$\times$ larger than prior efforts. This recipe provides a scalable path for transferring human manipulation and active-view priors to robot policies.

\item We develop AMP for active perception and mobile manipulation, enabling
single-operator scalable collection of demonstrations that jointly coordinate viewpoint,
bimanual manipulation, and mobility.
\end{itemize}

%% file: contents/related_work.tex
\subsection{Active Perception in Robot Manipulation}
% @shuaizhou
% Draft: Active Perception has been 
% /point: new best view prediction
% /via, sapave, egomi,activeumi,

Active perception has long been studied in robotics and computer vision
~\cite{bajcsy1988active,bajcsy2018revisiting}, with classical approaches often
formulated as Next-Best-View (NBV) planning
~\cite{connolly1985determination,bircher2016receding,breyer2022closed}.
These methods typically optimize viewpoint selection separately from downstream
manipulation. Recent learning-based methods instead couple viewpoint control with manipulation. ViA~\cite{xiong2025vision}, AV-ALOHA~\cite{chuang2025active},
and Eye, Robot~\cite{kerr2025eye} learn task-driven viewpoint control together
with manipulation, while EgoMI~\cite{yu2025egomi} and
ActiveUMI~\cite{zeng2025activeumi} exploit coordinated human viewpoint and
manipulation demonstrations. VLA-based approaches further model camera control within visuomotor policies: SaPaVe~\cite{liu2026sapave}
learns semantic camera control with decoupled camera and manipulation actions,
whereas ActiveMimic~\cite{lin2026activemimic} models camera and manipulation
motion jointly in a unified action space. Our work follows this direction while
explicitly grounding temporal camera motion in the VLA representation.

\subsection{Learning from Egocentric Human Videos}

Egocentric human videos provide a scalable source of diverse embodied
experience for robot learning beyond robot-collected data.
EgoMimic~\cite{kareer2025egomimic}, EgoBridge~\cite{punamiya2025egobridge},
and EgoScale~\cite{zheng2026egoscale} align human manipulation trajectories
with robot embodiments for policy transfer, while Kareer et
al.~\cite{kareer2025emergence} show that human-to-robot transfer can emerge
through joint training with sufficiently diverse robot pretraining. Importantly, egocentric videos encode more than
hand and wrist motion: human head-camera trajectories reflect
task-driven visual intent, as people actively reposition their viewpoints to
search for relevant objects, resolve occlusions, and support subsequent
manipulation. Recent works including
EgoMI~\cite{yu2025egomi} and
ActiveUMI~\cite{zeng2025activeumi} exploit this coordinated head--hand behavior
to learn active perception together with manipulation.
Our work builds on this perspective, using egocentric human data to transfer
both manipulation priors and task-driven viewpoint behaviors to robots.

\subsection{Spatial Intelligence for VLA}
%一些方案直接将深度信息作为模型输入，Point-VLA直接把点云注入进去, Stereo-VLA通过双目视觉形成立体感知, Spatial-VLA依靠深度估计专家估计深度信息并与rgb进行融合。Spatial Forcing首次通过特征对齐的方案让模型输入具有空间信息的特征, 基于这种思想，Lingbot-VLA和DualCoT-VLA进一步通过query来进行cot的方式感知深度。Motivated by ReconVLA，GEM-VLA和DepthVLA训练模型直接从2d输入中重建3d信息的能力。对于主动感知任务来说，关键问题不是对3D空间进行重建，而是理解Camera在一个 persistent scene下是如何变化来服务任务执行的。遵从第一性原理，our \method仅加入历史帧以建模camera pose的变化过程，并且训练模型根据history image and pose predict future的能力。
Existing methods enhance VLA spatial understanding through geometric inputs or auxiliary supervision.
PointVLA~\cite{li2026pointvla}, StereoVLA~\cite{deng2025stereovla}, and SpatialVLA~\cite{qu2025spatialvla} incorporate point clouds, stereo cues, and depth-derived spatial information, respectively.
Spatial Forcing~\cite{li2026spatial} first aligns visual representations with pretrained 3D features, while LingBot-VLA~\cite{wu2026pragmatic} and DualCoT-VLA~\cite{zhong2026dualcot} distill geometric knowledge into learnable queries.
Inspired by reconstruction-based VLAs~\cite{song2026reconvla}, GEM-VLA~\cite{zhao2026gem} and DepthVLA~\cite{yuan2025depthvla}
reconstruct depth from RGB observations as an auxiliary task.
For active perception tasks, the central challenge is not to reconstruct 3D geometry, but to understand how camera motion within a persistent scene supports task execution.
Accordingly, our \method~incorporates historical frames to model camera-pose evolution and predicts future observations conditioned on image and pose histories.

%% file: contents/methodology.tex
An overview of \method~is shown in Fig.~\ref{fig:head}.
In this section, we first formulate the active-perception manipulation problem
(Sec.~\ref{sec:method:problem formulation}), then introduce our robotic
platform and teleoperation setup
(Sec.~\ref{sec:method:robot setup}), followed by the model architecture
(Sec.~\ref{sec:method:model arch}) and training recipe
(Sec.~\ref{sec:method:training recipe}).

\subsection{Problem Formulation} \label{sec:method:problem formulation}

We formulate active-perception manipulation as learning a language-conditioned policy
$\pi_{\theta}$ that jointly controls manipulation and viewpoint motion.
At time $t$, given an observation $\mathbf{o}_{t}\in\mathcal{O}$ and a
language instruction $\ell\in\mathcal{L}$, the policy predicts a future
action chunk
$\mathbf{a}_{t:t+K-1}\in\mathcal{A}^{K}$, i.e.,
$\pi_{\theta}(\mathbf{o}_{t},\ell)\mapsto
\mathbf{a}_{t:t+K-1}$.
The observation $\mathbf{o}_{t}$ comprises visual observations and a state
context $\mathbf{s}_{t}$ containing non-visual robot information such as
proprioception and kinematics.
Each action $\mathbf{a}_{\tau}$ jointly specifies the Cartesian poses of the
left and right manipulation end effectors and the active camera, together with
the two gripper commands, all expressed in a shared robot-centric frame $\mathcal{F}_{U}$.

\subsection{AMP Setup} \label{sec:method:robot setup}

\begin{figure}[t]
    % \vspace{0.2cm}
    \centering
    \includegraphics[width=0.85\linewidth]{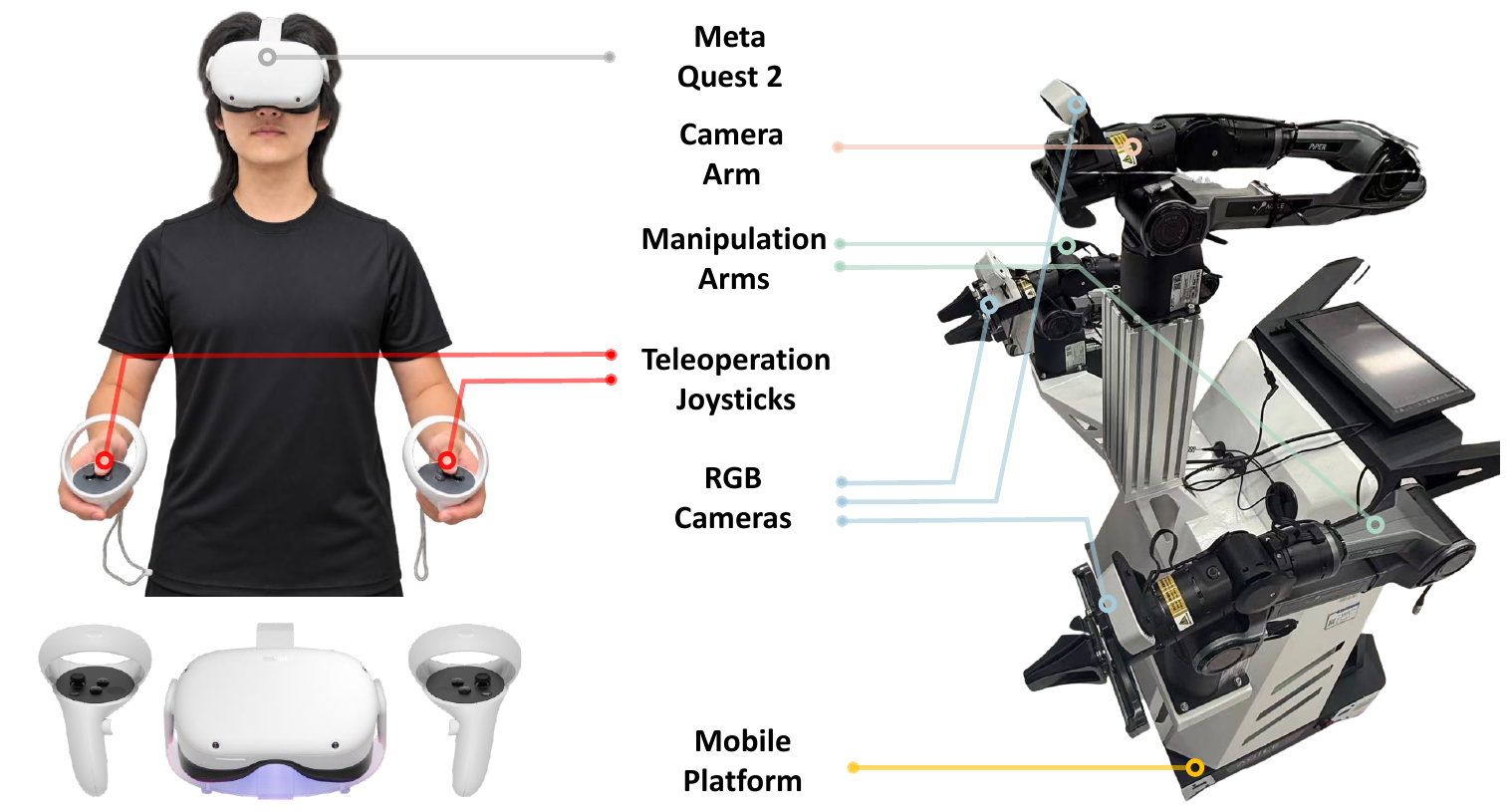}
    \caption{
AMP system overview. \textit{Left:} Single-operator Meta Quest 2 interface.
\textit{Right:} AMP triple-arm robotic platform.
    }
    \label{fig:robot}
    \vspace{-0.2cm}
\end{figure}

Our active-perception mobile-manipulation platform (AMP) is built on a modified AgileX Cobot-Magic, as shown in Fig.~\ref{fig:robot}.
The original platform consists of a differential-drive mobile base and two 6-DoF Piper arms, to which we add a third arm for active viewpoint control. The left and right arms perform bimanual manipulation and are each
equipped with an Orbbec DaBai DC1 camera near the end effector.
To enable active viewpoint control, we replace the original fixed front-facing
camera with an additional AgileX 6-DoF Piper arm carrying a third Orbbec DaBai DC1
camera at its end effector.
This design provides an independently actuated egocentric viewpoint while
preserving both manipulation arms and the mobile base for task execution.

\subsubsection{Teleoperation System}
To enable a single operator to jointly control the active viewpoint, bimanual manipulation, and mobile base, AMP uses a Meta Quest 2 headset and two handheld controllers as the teleoperation interface. 
At time $t$, we obtain the 6-DoF
poses $\mathbf{T}^{S}_{t}=(\mathbf{R}^{S}_{t},\mathbf{p}^{S}_{t})\in SE(3)$,
where $S\in\{H,L,R\}$ denotes the headset, left controller, and right
controller, respectively. The headset controls the middle arm carrying the
active camera \(C\), while the left and right controllers control the corresponding manipulation arms. 
We transform each tracked pose from the Quest tracking frame
to the robot control convention using a fixed axis transformation. Before
teleoperation, the three arms are initialized to predefined configurations that
approximately match the relative layout of the operator's head and hands. 
When the operator engages teleoperation by pressing both
controller grip buttons, we register the current tracked poses
$\mathbf{T}^{S}_{0}$ together with the current robot end-effector poses
$\mathbf{T}^{A}_{0}$ as reference configurations. Subsequent human motion is expressed relative to these references
as $\Delta\mathbf{p}^{S}_{t}=\mathbf{p}^{S}_{t}-\mathbf{p}^{S}_{0}$ and
$\Delta\mathbf{R}^{S}_{t}=\mathbf{R}^{S}_{t}
(\mathbf{R}^{S}_{0})^{\top}$. The corresponding robot target is then computed
as $\mathbf{p}^{A}_{t}=\mathbf{p}^{A}_{0}
+\lambda\Delta\mathbf{p}^{S}_{t}$ and
$\mathbf{R}^{A}_{t}=\Delta\mathbf{R}^{S}_{t}\mathbf{R}^{A}_{0}$,
with $\lambda=1$, and tracked through inverse kinematics.
Re-engaging
teleoperation re-anchors these references, allowing the operator to reposition
the headset and controllers without discontinuities.
The controller triggers continuously control the left and right grippers.
For mobile control, the left-controller X/Y buttons
rotate the base left/right, while the right-controller A/B buttons drive it
backward/forward, allowing the operator to navigate the workspace without
physically following the robot.  Thus, a single operator can coordinate camera motion,
bimanual manipulation, gripper control, and platform navigation within the same
interface.

\subsubsection{Data Collection}
We record synchronized demonstrations at 30\,Hz, including the three RGB
streams $\mathbf{I}_{t}=(I^{f}_{t},I^{l}_{t},I^{r}_{t})$, camera intrinsics
$(f_x,f_y,c_x,c_y)$, distortion parameters, image resolution, and joint feedback and control commands for all three arms. To obtain a representation that remains local to the robot as the mobile base moves, we express the realized poses of both manipulation end effectors and the active camera in a shared robot-centric frame $\mathcal{F}_{U}$ rigidly attached to the mobile platform. Specifically, $\mathcal{F}_{U}$ shares the orientation of the middle-arm base frame, with its origin shifted $0.25$\,m downward along the middle-arm $z$ axis. Thus, $\mathcal{F}_{U}$ moves together with the robot rather than remaining fixed at the platform's initial world pose. For $X\in\{L,R,C\}$, corresponding to the left end effector, right end effector, and active camera, respectively, we represent its pose as ${}^{U}\boldsymbol{\xi}^{X}_{t}=[{}^{U}\mathbf{p}^{X}_{t},{}^{U}\mathbf{q}^{X}_{t}]\in\mathbb{R}^{7}$, where ${}^{U}\mathbf{p}^{X}_{t}\in\mathbb{R}^{3}$ denotes position in $\mathcal{F}_{U}$ and ${}^{U}\mathbf{q}^{X}_{t}$ is an $xyzw$-ordered unit quaternion. Together with the two gripper positions, this gives the 23-D robot-centric state $\mathbf{s}_{t}=[{}^{U}\boldsymbol{\xi}^{L}_{t},{}^{U}\boldsymbol{\xi}^{R}_{t},{}^{U}\boldsymbol{\xi}^{C}_{t},\gamma^{L}_{t},\gamma^{R}_{t}]$. We use the corresponding realized Cartesian trajectories and teleoperated gripper commands as pose-space action targets in the same representation. We separately record the planar motion of the mobile base as $\mathbf{b}_{t}=(v_{t},\omega_{t})$, where $v_{t}$ and $\omega_{t}$ denote the forward linear and yaw angular velocities, respectively.

\subsection{Model Architecture and Objectives} \label{sec:method:model arch}

\begin{figure}[t]
    % \vspace{0.2cm}
    \centering
    \includegraphics[width=0.9\linewidth]{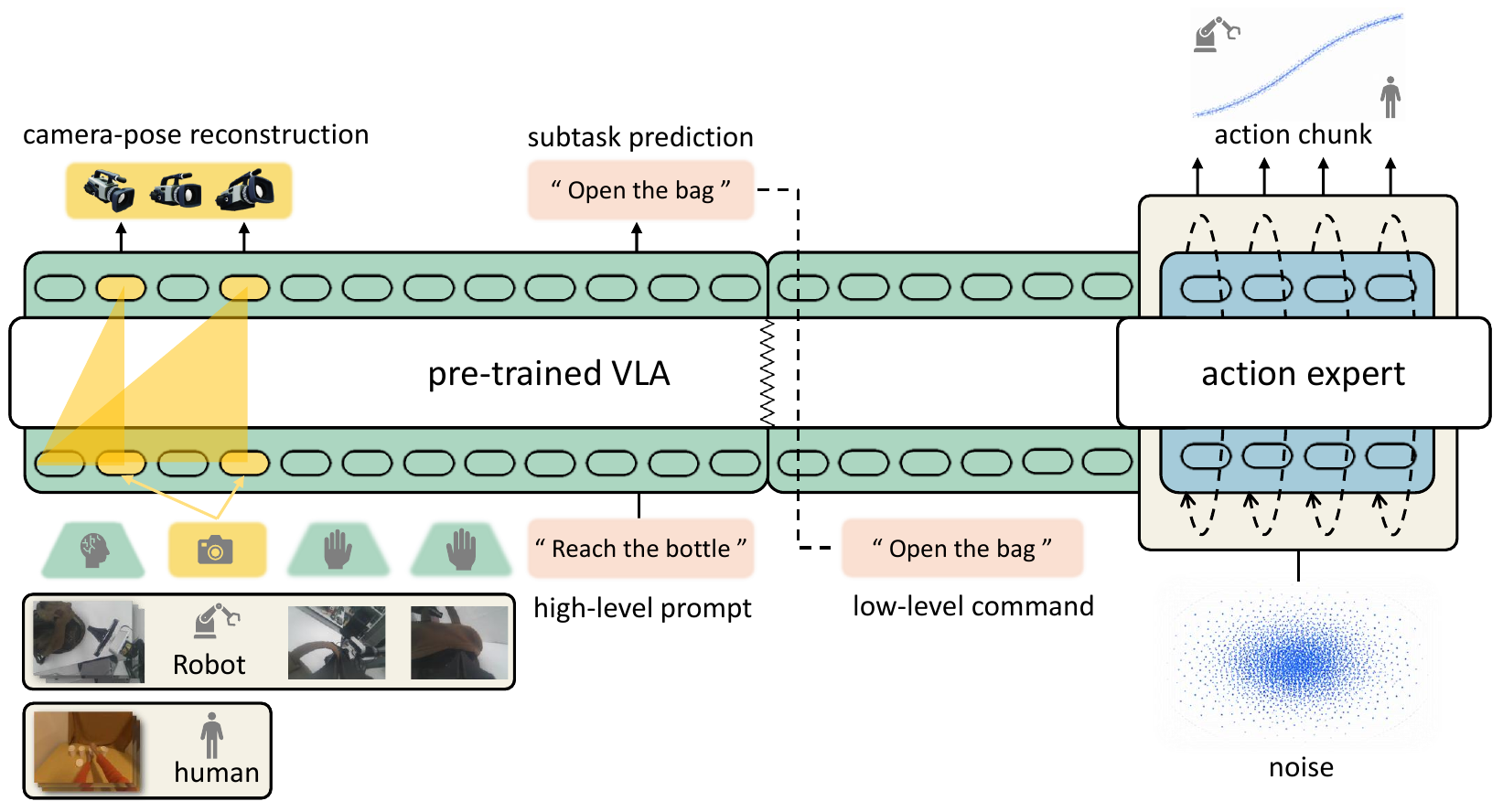}
    \caption{
Model overview. A pretrained VLA backbone encodes visual, language, and state
context. Pose-grounded camera tokens support camera-pose reconstruction, while the
VLM predicts semantic outputs such as subtasks and a flow-matching action
expert generates manipulation and active-camera actions. The architecture is
shared across human and robot training data.
    }
    \label{fig:model}
    \vspace{-0.2cm}
\end{figure}

% Active perception requires a policy not only to decide where to look, but also
% to understand how successive visual observations are related as the viewpoint changes. This requires the model to jointly reason about temporal visual
% observations, camera motion, and manipulation when predicting future actions. As illustrated in Fig.~\ref{fig:model}, we build on
% $\pi_{0.5}$~\cite{intelligence2025pi_} and augment its visual context with pose-grounded camera tokens inspired by
% Cambrian-P~\cite{yang2026cambrianp}, while treating active-camera motion as part of the action space.
Active perception requires a policy to relate visual observations across
changing viewpoints while jointly reasoning about camera motion and
manipulation. As shown in Fig.~\ref{fig:model}, we build on
$\pi_{0.5}$~\cite{intelligence2025pi_} and augment its visual context with pose-grounded camera tokens inspired by Cambrian-P~\cite{yang2026cambrianp}, while treating camera motion as part of the action space.

\subsubsection{$\pi_{0.5}$ Backbone}

We adopt $\pi_{0.5}$~\cite{intelligence2025pi_}
as our base VLA architecture. $\pi_{0.5}$ consists of a pretrained PaliGemma~\cite{beyer2024paligemma}
vision--language model (VLM) and a lightweight transformer-based action expert.
The VLM encodes multi-view visual
observations, language instructions, and tokenized proprioceptive states to
provide context for action generation.
Conditioned on this context, the action expert predicts continuous action
chunks using flow matching. Given a clean action chunk $\mathbf{a}$,
Gaussian noise $\boldsymbol{\epsilon}\sim\mathcal{N}(0,I)$, and flow time
$\tau\in[0,1]$, we construct
$\mathbf{x}_{\tau}
=\tau\boldsymbol{\epsilon}+(1-\tau)\mathbf{a}$ with target velocity
$\mathbf{u}_{\tau}=\boldsymbol{\epsilon}-\mathbf{a}$.
The action expert predicts a conditional velocity field
$\mathbf{v}_{\theta}(\mathbf{x}_{\tau},\tau,\mathbf{c})$, where
$\mathbf{c}$ denotes the VLM context, and is trained with
$ \mathcal{L}_{\mathrm{flow}}
    =
    \mathbb{E}
    \left[
    \left\|
    \mathbf{v}_{\theta}(\mathbf{x}_{\tau},\tau,\mathbf{c})
    - \mathbf{u}_{\tau}
    \right\|_2^2
    \right].
$
In addition to flow-matching supervision, we retain the
autoregressive token-prediction objectives of $\pi_{0.5}$.
At the high level, the VLM predicts the current subtask from the episode-level
instruction and visual-state context, supervised with a standard next-token
cross-entropy loss $\mathcal{L}_{\mathrm{subtask}}$.
At the low level, the current subtask serves as the action-conditioning
instruction. FAST~\cite{pertsch2025fast} converts
the future continuous action trajectory into discrete action tokens, which are
autoregressively predicted with next-token cross-entropy loss
$\mathcal{L}_{\mathrm{FAST}}$.
The same subtask-conditioned context also conditions the continuous
flow-matching action expert.

\subsubsection{Camera Tokens}

To provide spatial context across changing viewpoints, we augment the $\pi_{0.5}$ input with historical observations that retain previously observed scene information. We associate each frame with a learnable  camera token~\cite{yang2026cambrianp}, supervised to predict its camera
pose and field of view, encouraging the model to relate observations across viewpoints.
We use the current front-camera observation together with three past observations sampled at 16-frame intervals, denoted by $\mathcal{H}^{f}_{t} = (I^{f}_{t-48}, I^{f}_{t-32}, I^{f}_{t-16}, I^{f}_{t})$.
Each frame is encoded by the shared vision encoder, and a learnable camera
token is appended after its visual tokens. Specifically, we use two learnable embeddings
$\mathbf{c}_{\mathrm{first}}$ and $\mathbf{c}_{\mathrm{rest}}$ for the
earliest valid frame and subsequent frames, respectively.
The visual-language prefix is therefore organized as
$[\mathbf{V}^{f}_{t-48},\mathbf{c}_{1},\ldots,
\mathbf{V}^{f}_{t},\mathbf{c}_{4},
\mathbf{V}^{l}_{t},\mathbf{V}^{r}_{t},
\mathbf{E}(\ell,\mathbf{s}_{t})]$,
where $\mathbf{V}$ denotes visual tokens and
$\mathbf{E}(\ell,\mathbf{s}_{t})$ denotes the tokenized language and
proprioceptive-state context. We apply block-causal attention over the front-camera history: each temporal
block can attend to itself and preceding history, while its camera token
cannot access future frames. The subsequent wrist, language, and action
tokens can attend to all valid camera-history blocks.

The hidden state of each camera token is linearly projected
to a VGGT-style camera head~\cite{wang2025vggt}, which predicts a 9-D camera
representation consisting of 3D translation $(x,y,z)$, quaternion orientation
$(q_x,q_y,q_z,q_w)$, and vertical and horizontal fields of view
$(\mathrm{fov}_y,\mathrm{fov}_x)$.
For camera tokens with valid geometric targets, we supervise the camera head
with
\begin{equation}
\label{eq:camera_loss}
\mathcal{L}_{\mathrm{cam}}
=
\mathcal{L}_{\mathrm{trans}}
+
\mathcal{L}_{\mathrm{rot}}
+
\lambda_{\mathrm{fov}}\mathcal{L}_{\mathrm{fov}},
\end{equation}
where the three terms are component-wise $\ell_1$ losses on translation,
quaternion orientation, and field of view, respectively, and we use
$\lambda_{\mathrm{fov}}=0.5$. This auxiliary head geometrically grounds the camera-token
representation and is removed during action inference. The resulting prefix
conditions the original $\pi_{0.5}$ action expert, which jointly generates
manipulation and head camera motions.

\subsection{Training Recipe} \label{sec:method:training recipe}
Starting from the pretrained $\pi_{0.5}$ checkpoint~\cite{intelligence2025pi_}, which
provides broad visual-semantic and robot-control priors from large-scale
pretraining, we train our policy in two stages.
We first perform human--robot mid-training, leveraging diverse egocentric human demonstrations to acquire manipulation and active-view priors while
interleaving robot demonstrations to maintain robot-aligned action learning.
We then perform task-specific post-training on our teleoperation
data to adapt the policy to the target embodiment and
tasks.
\subsubsection{Human-Robot Co-training} Egocentric human data provide scalable supervision for transferable manipulation priors~\cite{zheng2026egoscale}.
Recent work further shows that
human-to-robot transfer can emerge through human--robot co-training with a
sufficiently diverse pretrained VLA~\cite{kareer2025emergence}. Meanwhile,
continuous action adaptation can interfere with semantic knowledge inherited
from the pretrained VLM~\cite{driess2026knowledge}.
Motivated by these observations, we jointly train on human and robot
demonstrations, leveraging human data to acquire transferable semantic,
manipulation, and active-view priors while retaining robot-aligned
action learning. 
Human and robot demonstrations are sampled at a
balanced 1:1 ratio.
For human demonstrations, head-camera poses provide
active-view supervision and wrist trajectories provide manipulation
supervision. 
Following~\cite{lin2026activemimic}, we re-express the human
camera and wrist trajectories in the camera frame of the earliest valid
observation in each temporal window, while robot trajectories remain in their
fixed robot reference frames.
We normalize states and action trajectories separately for each data source and mask unavailable action dimensions and camera-pose targets.

The human data are drawn from the large-scale and diverse
egocentric datasets EgoLive~\cite{li2026egolive}
and EgoVerse~\cite{punamiya2026egoverse}, which provide
rich first-person manipulation experience together with head-camera motion.
We further curate these datasets by retaining indoor household-manipulation
clips with valid camera trajectories and the task/action supervision required
by our training pipeline, while prioritizing activities semantically related
to our downstream settings.
The robot portion combines diverse demonstrations from
AgiBot World~\cite{bu2025agibot,agibotworld2026},
RoboCOIN~\cite{wu2025robocoin}, and our AMP,
covering diverse robot embodiments
and control distributions. 

The mixed-batch training objective is
$\mathcal{L}_{\mathrm{co}}
=\mathcal{L}^{H}_{\mathrm{subtask}}
+\lambda_H(\mathcal{L}^{H}_{\mathrm{FAST}}+\mathcal{L}^{H}_{\mathrm{flow}})
+\mathcal{L}^{R}_{\mathrm{FAST}}
+\mathcal{L}^{R}_{\mathrm{flow}}
+\lambda_C\mathcal{L}_{\mathrm{cam}}$.
Here, $H$ and $R$ denote human and robot examples, and each term is evaluated only when the corresponding supervision is available in the data. High-level human samples supervise $\mathcal{L}^{H}_{\mathrm{subtask}}$, while low-level human and robot
samples supervise their FAST and flow objectives. We use
$\lambda_H=\lambda_C=0.2$.

\subsubsection{Post-training with Teleoperation Data}
After mid-training, we perform task-specific post-training on the
AMP teleoperation demonstrations collected with the system described
above. We keep the model architecture and action representation unchanged and
optimize
$\mathcal{L}_{\mathrm{post}}
=
\mathcal{L}_{\mathrm{flow}}
+
\lambda_C\mathcal{L}_{\mathrm{cam}}$,
with $\lambda_C=0.2$.
Unlike in human--robot co-training, we do not apply the autoregressive subtask or
FAST objectives, since the post-training data directly provide aligned robot
action and active-camera trajectories in the target embodiment.
This stage specializes the co-trained policy to the AMP
and downstream manipulation tasks.

\begin{figure*}[t]
    % \vspace{0.2cm}
    \centering
    \includegraphics[width=0.8\textwidth]{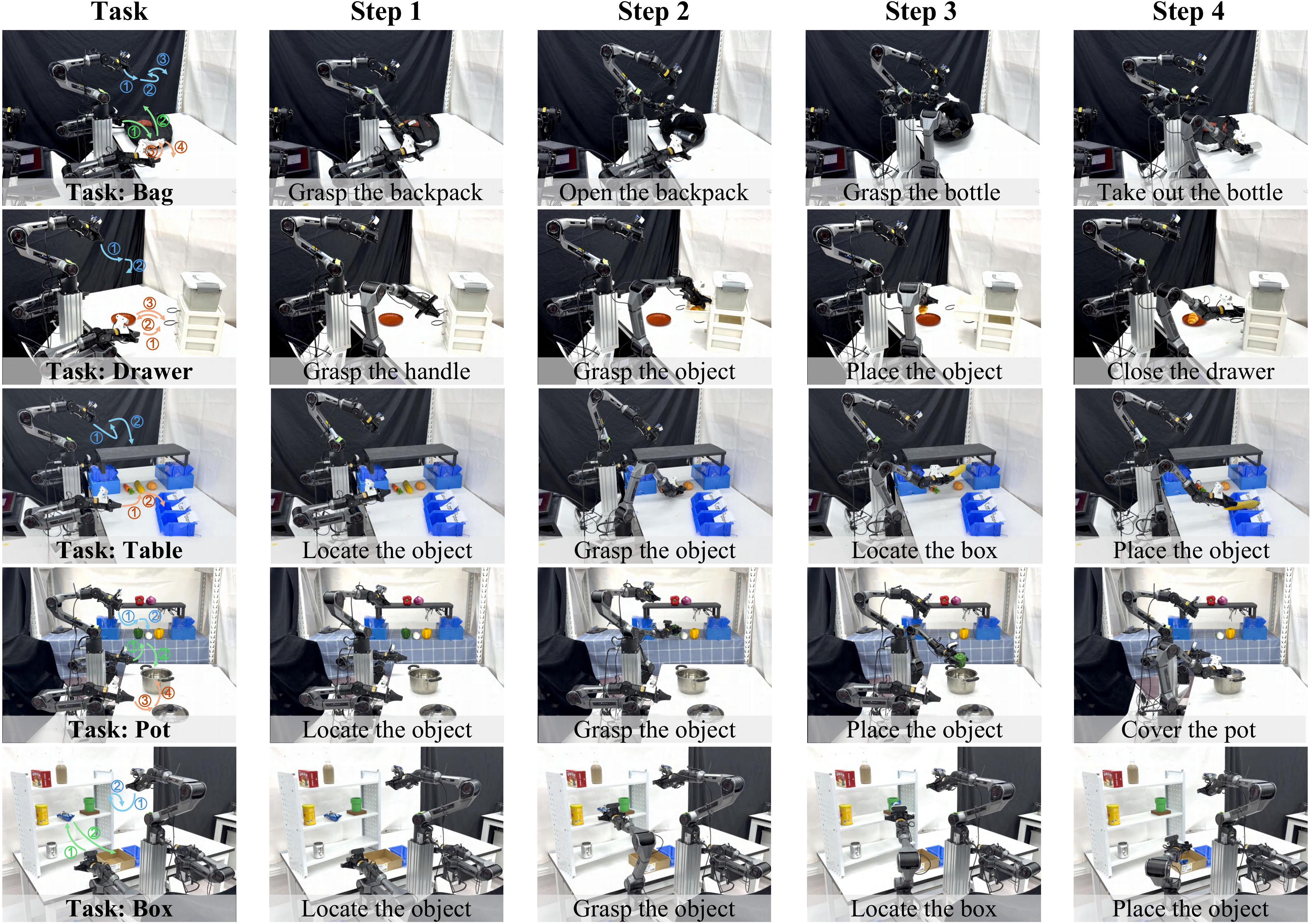}
    \caption{
    Overview of the five real-world active-perception task families, grouped
    into two complementary settings:
    \textit{Occlusion-Driven Manipulation} (Bag, Drawer, and Table), where
    task-relevant objects are initially hidden or outside the default view,
    and \textit{Active Search and Localization} (Pot and Box), where the robot
    must actively search across workspace regions and localize instructed
    objects or targets.
    }
    \label{fig:tasks}
    \vspace{-0.2cm}
\end{figure*}

%% file: contents/experiments.tex
We train our models on NVIDIA H100 GPUs and deploy them on AMP for real-world
evaluation, focusing on tasks in which successful execution requires the robot
to actively acquire task-relevant visual information during manipulation.
Our evaluation addresses four complementary questions:
\begin{itemize}[leftmargin=*,nosep]
    \item \textbf{Q1: Active-perception performance.}
    How well does our policy perform on real-world manipulation tasks that
    require active viewpoint control?

    \item \textbf{Q2: Human--robot transfer.}
    Does human--robot mid-training improve downstream robot performance?

    \item \textbf{Q3: Model design.}
    How do historical observations and pose-supervised camera tokens contribute
    to active-perception performance?

    \item \textbf{Q4: Execution efficiency.}
    Can our policy meet the requirements of real-time deployment?
\end{itemize}
We first introduce the real-world task suite and evaluation protocol.
For a reproducible comparison, we evaluate against $\pi_{0.5}$, a state-of-the-art VLA for real-world robotic manipulation, as recent active-perception VLAs do not publicly release checkpoints and complete training recipes for direct evaluation. We then study the effects of human--robot mid-training and our
camera-aware VLA model design through controlled ablations, and analyze corresponding execution efficiency. Finally, we demonstrate AMP on a
single-operator mobile-manipulation task that combines active viewpoint control,
bimanual manipulation, and base motion.

\subsection{Task Suite and Evaluation Protocol}

Active perception is particularly important when task-relevant information is
initially occluded or outside the robot's field of view, or when the robot must
search a larger workspace to localize relevant objects or targets. To cover
these two settings, we design five real-world task families, illustrated in
Fig.~\ref{fig:tasks}, that induce complementary forms of
viewpoint demand.

% \textbf{Occlusion-Driven Manipulation.}
\begin{itemize}[leftmargin=*,nosep]
    \item \textbf{Bag.}
    The robot retrieves a bottle from inside a backpack and places it on the
    table. The bottle is initially occluded, requiring the robot to inspect the
    backpack interior while opening it; success of this task requires the bottle to be
    stably placed on the table.

    \item \textbf{Drawer.}
    The robot opens the upper or middle drawer, retrieves the instructed
    object, places it on the plate, and closes the drawer. The target object becomes
    visible only after opening; success of this task requires the object on the plate and
    the drawer closed.

    \item \textbf{Table.}
    The robot retrieves an instructed object underneath the table and places it
    into the specified labeled bin. The target is initially outside the frontal
    view, requiring downward viewpoint adjustment; success of this task requires the object
    in the instructed bin.
\end{itemize}

% \textbf{Active Search and Localization.}
\begin{itemize}[leftmargin=*,nosep]
    \item \textbf{Pot.}
    The robot retrieves the instructed vegetable from the shelf, places it in
    the pot, and covers the pot with the lid. The robot must shift its viewpoint
    across the shelf, pot, and lid; success requires the correct vegetable in
    the pot with the lid stably placed.

    \item \textbf{Box.}
    The robot retrieves an instructed object from the shelf and places it into
    the left or right target box according to the language instruction.
    Different shelf heights require vertical viewpoint search; success requires
    the object in the instructed target box.
\end{itemize}

Together, these tasks span container and articulated occlusion, below-table
perception, multi-stage viewpoint transitions, and large vertical workspace
variation. For each task family, we use 150 demonstrations for post-training
and evaluate each method over 20 real-world rollouts under matched conditions.
A rollout succeeds only if all required stages and the task-specific final-state
criterion are satisfied. We report strict task success rate (SR) and task
progress (TP). TP sums the credits for completed stages in
Table~\ref{tab:tp-scoring} and is averaged over all rollouts for each task.

% \begin{table*}[t]
%     \small
%     \centering
%     \caption{Task progress scoring criteria. Parentheses indicate
%     the incremental credit awarded for completing each stage.}
%     \label{tab:tp-scoring}
%     \small
%     \setlength{\tabcolsep}{4pt}
%     \renewcommand{\arraystretch}{1.15}
%     \begin{tabular*}{\textwidth}{@{\extracolsep{\fill}}lllll@{}}
%         \toprule
%         Task & Stage 1 & Stage 2 & Stage 3 & Stage 4 \\
%         \midrule
%         \textbf{Bag}
%         & Grasp backpack (0.3)
%         & Grasp bottle (0.3)
%         & Place bottle on table (0.4)
%         & --- \\
%         \textbf{Drawer}
%         & Open drawer (0.3)
%         & Grasp target object (0.3)
%         & Place object on plate (0.2)
%         & Close drawer (0.2) \\
%         \textbf{Under}
%         & Grasp target object (0.5)
%         & Place object in target bin (0.5)
%         & ---
%         & --- \\
%         \textbf{Pot}
%         & Grasp target vegetable (0.4)
%         & Place vegetable in pot (0.2)
%         & Grasp lid (0.2)
%         & Cover pot (0.2) \\
%         \textbf{Box}
%         & Grasp target object (0.5)
%         & Place object in target box (0.5)
%         & ---
%         & --- \\
%         \bottomrule
%     \end{tabular*}
% \end{table*}

\begin{table*}[t]
    \centering
    \caption{Task progress scoring criteria. Parentheses indicate stage credit; dashes denote inapplicable stages.}
    \label{tab:tp-scoring}
    \footnotesize
    \setlength{\tabcolsep}{3pt}
    \renewcommand{\arraystretch}{0.95}
    \begin{tabular}{lllll}
        \toprule
        Task & Stage 1 & Stage 2 & Stage 3 & Stage 4 \\
        \midrule
        \textbf{Bag}
        & Grasp backpack (0.3)
        & Grasp bottle (0.3)
        & Place bottle on table (0.4)
        & --- \\
        \textbf{Drawer}
        & Open drawer (0.3)
        & Grasp target object (0.3)
        & Place object on plate (0.2)
        & Close drawer (0.2) \\
        \textbf{Table}
        & Grasp target object (0.5)
        & Place object in target bin (0.5)
        & ---
        & --- \\
        \textbf{Pot}
        & Grasp target vegetable (0.4)
        & Place vegetable in pot (0.2)
        & Grasp lid (0.2)
        & Cover pot (0.2) \\
        \textbf{Box}
        & Grasp target object (0.5)
        & Place object in target box (0.5)
        & ---
        & --- \\
        \bottomrule
    \end{tabular}
    \vspace{-0.4cm}
\end{table*}

\subsection{Main Experiments (\textbf{Q1})}

\begin{figure}[t]
    \centering
    \includegraphics[width=0.75\linewidth]{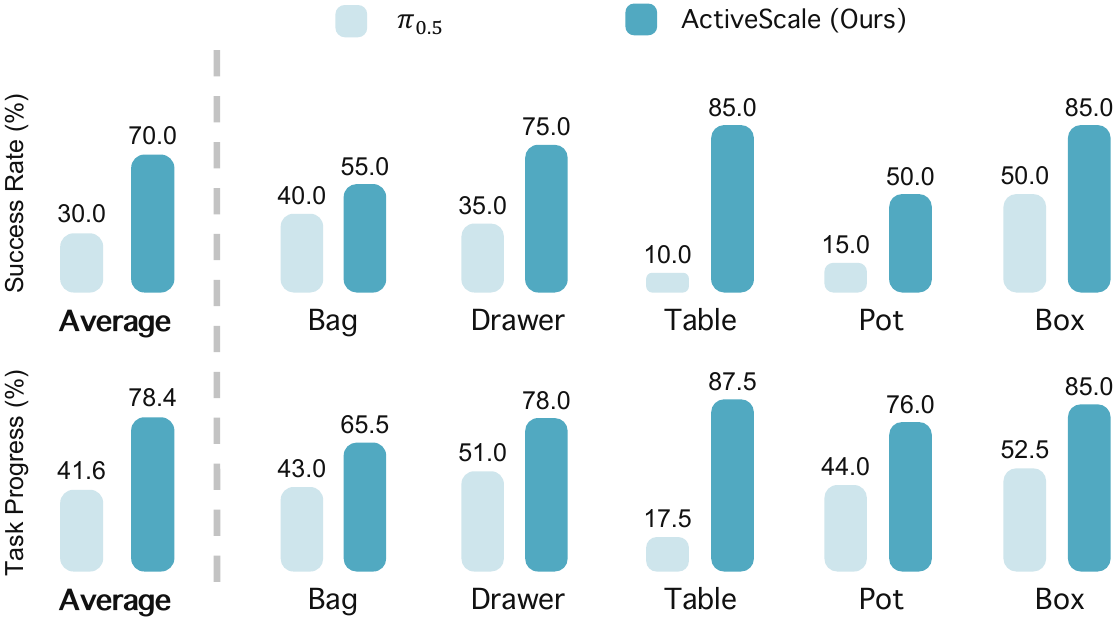}
    \caption{Comparison with $\pi_{0.5}$ across five manipulation tasks.}
    \label{fig:exp-pi}
    \vspace{-0.1cm}
\end{figure}

We compare the complete two-stage \method~pipeline against the $\pi_{0.5}$
baseline under the same downstream post-training and evaluation protocol.
The baseline is directly post-trained from the pretrained $\pi_{0.5}$
checkpoint on the task-specific demonstrations, whereas \method~undergoes
human--robot mid-training before post-training on the same data. As shown in Fig.~\ref{fig:exp-pi},
\method~achieves higher SR and TP across all five task families, increasing
mean SR from 30.0\% to 70.0\% and the mean TP from 41.6\% to 78.4\%.
These gains indicate that the proposed model is substantially more effective
at completing manipulation tasks in which useful visual information must be
actively acquired during execution. In particular, the improvements are most
pronounced when the target is initially hidden or outside the default view.
On \textit{Table}, $\pi_{0.5}$ can stall without initiating a grasp,
whereas \method~redirects its viewpoint, localizes the target, and proceeds
with manipulation. Similarly, on \textit{Bag}, \method~actively inspects the backpack
interior while opening it, enabling subsequent object retrieval. Overall,
these results show that \method~more effectively coordinates viewpoint
adjustment with manipulation and exploits newly acquired observations to
advance task execution.

\subsection{Ablation Studies}

We next isolate the contributions of our training recipe and model design, corresponding to Q2 and Q3. Unless otherwise specified, all variants use the same task-specific post-training data and real-world evaluation protocol.

\begin{figure}[t]
    \centering
    \includegraphics[width=0.75\linewidth]{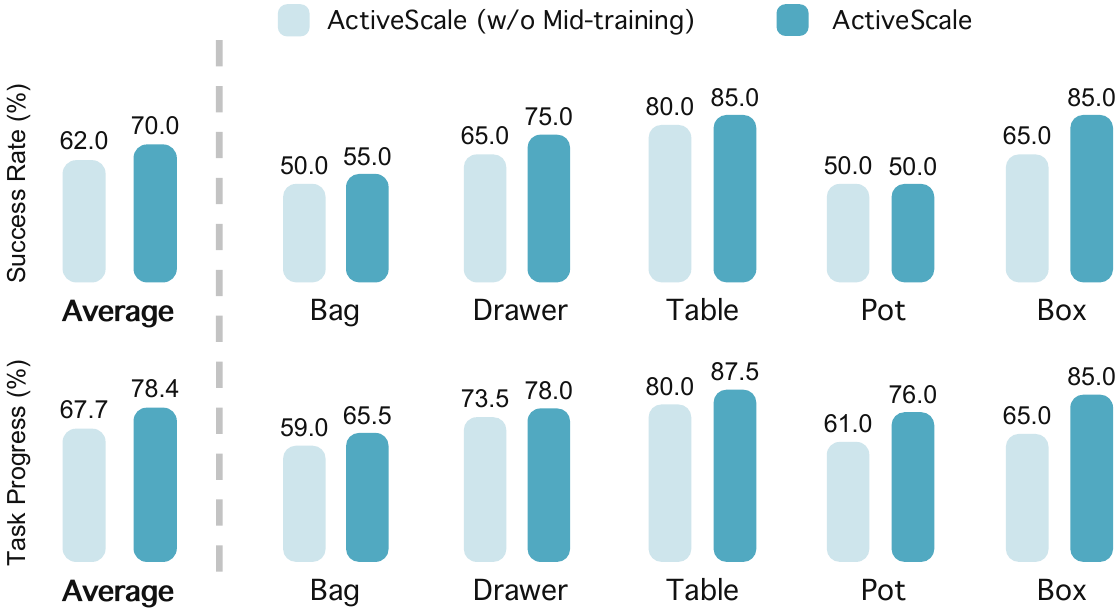}
    \caption{Ablation study on the effect of mid-training.}
    \label{fig:exp-mid}
    \vspace{-0.1cm}
\end{figure}

\begin{figure*}[t]
    % \vspace{0.2cm}
    \centering
    \includegraphics[width=0.8\textwidth]{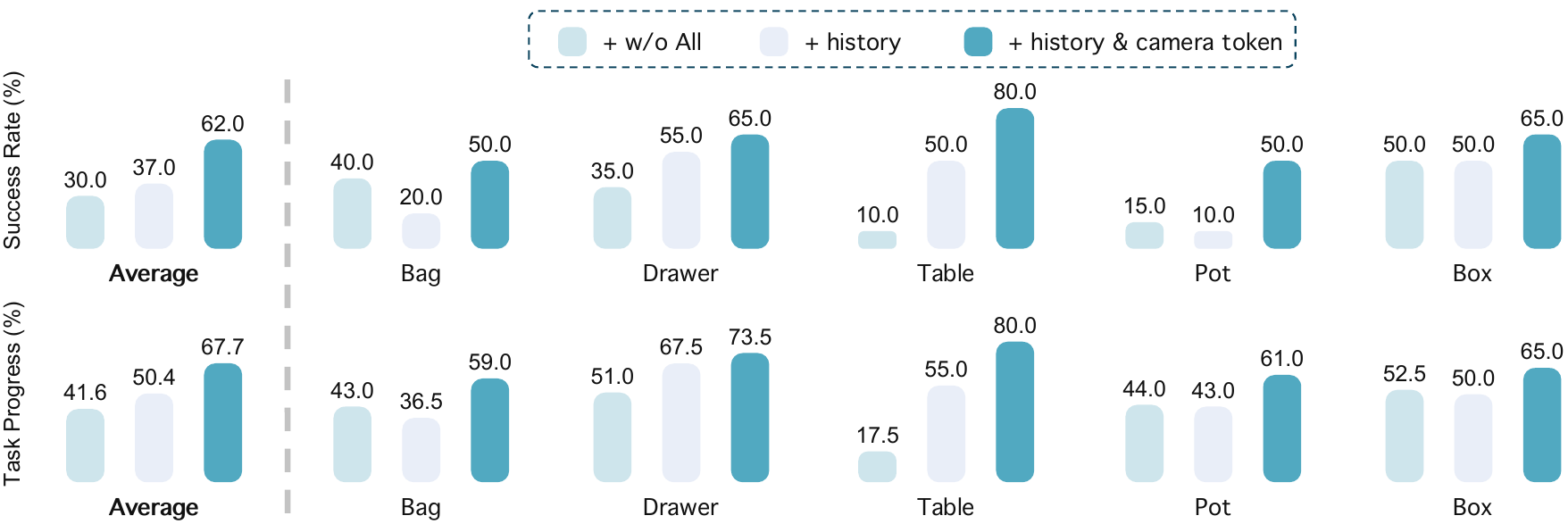}
    \caption{Ablation study on history and camera token settings.}
    \label{fig:exp-abla}
    \vspace{-0.4cm}
\end{figure*}

\subsubsection{Human--Robot Mid-Training (\textbf{Q2})}
We isolate the effect of human--robot mid-training by comparing two variants
of our model with the same architecture, task-specific demonstrations, and
post-training protocol. The variant without mid-training is directly
post-trained from the pretrained $\pi_{0.5}$ checkpoint, whereas the full
model first undergoes human--robot mid-training and is then post-trained on
the same downstream data. As shown in Fig.~\ref{fig:exp-mid}, human--robot
mid-training improves overall downstream performance, although the magnitude
of the gain varies across tasks. On \textit{Pot}, for example, TP improves while SR remains unchanged. Qualitatively, the mid-trained policy produces
better-aligned lid placements, whereas the variant without mid-training can
misalign the lid and hesitate before release. This suggests that mid-training
can improve intermediate execution quality even when the improvement is not
fully reflected by binary task success. Overall, these results support our
two-stage training recipe of human--robot mid-training followed by
task-specific post-training.

\subsubsection{Temporal History and Camera Tokens (\textbf{Q3})}
To isolate the contributions of temporal history and pose-grounded camera
tokens from the effect of mid-training, we compare three variants using
task-specific post-training only: original $\pi_{0.5}$, a history-only
variant, and our full temporal representation with historical observations and
pose-grounded camera tokens. As shown in Fig.~\ref{fig:exp-abla}, adding
historical observations alone has a task-dependent effect, improving
performance on \textit{Drawer} and \textit{Table} but not consistently across all task families.
On \textit{Bag}, for example, the history-only policy sometimes fails to open the
backpack and performs worse than the single-frame base policy. These mixed
results suggest that retaining past observations can provide useful temporal
context, but history alone is insufficient to reliably interpret observations
acquired from changing viewpoints. In contrast, adding pose-grounded camera
tokens improves both SR and TP across all five task families relative to the
history-only variant and recovers the performance drops observed on \textit{Bag} and
\textit{Pot}. This consistent improvement supports explicitly grounding temporal
observations with camera motion, helping the policy relate visual information
acquired from different viewpoints.

\subsection{Execution Efficiency (\textbf{Q4})}

To quantify our model’s inference efficiency, we adopt the action generation throughput metric used in OpenVLA-OFT~\cite{kim2025fine}. With RTC~\cite{black2026real} enabled, \method~generates 50-action chunks with a throughput of 264.5 Hz on a single RTX 4090. Asynchronous execution overlaps inference with ongoing actions to eliminate waiting delays between action chunks, supporting continuous execution at the required control frequency of 30 Hz.

\subsection{AMP Teleoperation Demonstration}
\begin{figure}[t]
    \centering
    \includegraphics[width=0.55\linewidth]{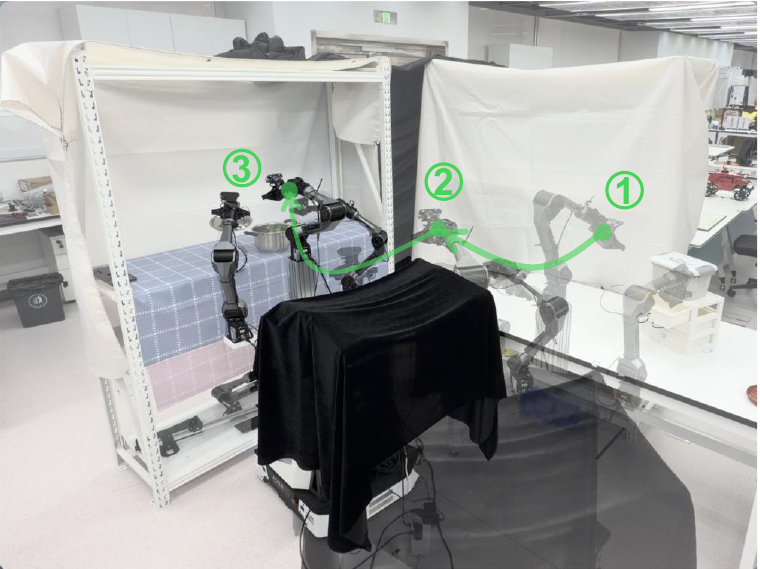}
    \caption{
Single-operator AMP teleoperation demonstration combining active perception
and mobile manipulation. See the supplementary video for the full sequence.
    }
    \label{fig:mobile}
    \vspace{-0.2cm}
\end{figure}

To demonstrate the capability of AMP as a data-collection platform, we
teleoperate a multi-stage mobile-manipulation task with a single operator, as
shown in Fig.~\ref{fig:mobile}. The robot first actively adjusts its viewpoint
to inspect a drawer and retrieve a vegetable, then redirects the camera to
locate the cooking area before navigating the mobile base toward it. After
reaching the destination, the operator completes the manipulation sequence by
placing the vegetable into the pot and covering it with the lid. This
demonstration shows that AMP enables a single operator to seamlessly coordinate
active viewpoint control, bimanual manipulation, and mobile-base motion within
one teleoperation interface.

%% file: contents/conclusions.tex
% We presented \method, a framework for active-perception robot learning across
% model, data, and hardware. On the model side, \method~augments a VLA with
% temporal visual observations and pose-grounded camera tokens to explicitly
% reason about viewpoint changes across time. On the data side, we introduce a
% human--robot mid-training recipe that leverages egocentric human video together
% with robot demonstrations to improve downstream active-perception learning.
% On the hardware side, we develop AMP, a single-operator teleoperation platform
% that jointly supports active viewpoint control, bimanual manipulation, and
% mobile-base motion. Across five real-world active-perception task families,
% \method~substantially outperforms directly post-trained $\pi_{0.5}$, improving
% mean task success from 30.0\% to 70.0\%, while our ablations demonstrate the
% benefits of both human--robot mid-training and pose-grounded temporal modeling.
% Our current experiments focus on mid-training as the primary mechanism for
% leveraging egocentric data; future work can investigate how active-perception
% capabilities scale with substantially larger and more diverse datasets,
% including active-perception-oriented pretraining. In addition, while AMP
% demonstrates single-operator data collection for mobile manipulation with
% active perception, our policy evaluation currently focuses on the five
% task families studied here. Scaling data collection and systematically
% evaluating learned policies on mobile active-perception tasks therefore

We presented \method, a framework for active-perception robot learning across
model, data, and hardware. \method~augments a VLA with temporal observations
and pose-grounded camera tokens, introduces a human--robot mid-training recipe
using egocentric human and robot data, and develops AMP, a single-operator
platform for active viewpoint control, bimanual manipulation, and mobile-base
motion. Across five real-world active-perception task families, \method~
outperforms directly post-trained $\pi_{0.5}$, improving mean task success from
30.0\% to 70.0\%, while ablations validate the benefits of human--robot
mid-training and pose-grounded temporal modeling. 
%投稿版本不讲limitations，少讲future work
% Our current study focuses on mid-training as the primary mechanism for leveraging egocentric data; future work can explore active-perception pretraining with larger and more diverse datasets. 
% While AMP demonstrates single-operator mobile manipulation with active perception, our policy evaluation remains limited to the five task families studied here. 
Future work will scale AMP data collection and
systematically evaluate learned policies on mobile active-perception tasks.